\documentclass[final]{cvpr}

\usepackage{times}
\usepackage{epsfig}
\usepackage{graphicx}
\usepackage{amsmath}
\usepackage{amssymb}

\usepackage{subcaption}

\usepackage[pagebackref=true,breaklinks=true,colorlinks,bookmarks=false]{hyperref}

\newcommand{\tbf}{\textcolor[rgb]{0.0,0.0,0.0}}

\begin{document}
\title{\vspace{-41pt}\tbf{Towards Self-Supervision for Video Identification\\of Individual Holstein-Friesian Cattle: The \textsl{Cows2021} Dataset}}  
\author{Jing Gao$^1$ \hspace{5pt} Tilo Burghardt$^1$ \hspace{5pt} William Andrew$^{1,2}$ \hspace{5pt} Andrew W. Dowsey$^2$ \hspace{5pt} Neill W. Campbell$^1$ \\ \ \\ $^1$Department of Computer Science \hspace{50pt} $^2$Bristol Veterinary School\\ \hspace{20pt}University of Bristol \hspace{100pt} University of Bristol\\ \hspace{20pt}Bristol, United Kingdom \hspace{80pt} Bristol, United Kingdom}
\vspace{-11pt}
\maketitle

\begin{abstract}
\tbf{In this paper we publish the largest identity-annotated Holstein-Friesian cattle dataset (\textsl{Cows2021}) and a first self-supervision framework for video identification of individual animals. The dataset contains $10,402$ RGB images with labels for localisation and identity as well as $301$ videos from the same herd. The data shows top-down in-barn imagery, which captures the breed's individually distinctive black and white coat pattern. Motivated by the labelling burden involved in constructing visual cattle identification systems, we propose exploiting the temporal coat pattern appearance across videos as a self-supervision signal for animal identity learning. Using an individual-agnostic cattle detector that yields oriented bounding-boxes, rotation-normalised tracklets of individuals are formed via tracking-by-detection and enriched via augmentations. This produces a `positive' sample set per tracklet, which is paired against a `negative' set sampled from random cattle of other videos. Frame-triplet contrastive learning is then employed to construct a metric latent space. The fitting of a Gaussian Mixture Model to this space yields a cattle identity classifier. Results show an accuracy of {Top-1: $57.0\%$ and Top-4: $76.9\%$} and an Adjusted Rand Index: $0.53$ compared to the ground truth. Whilst supervised training surpasses this benchmark by a large margin, we conclude that self-supervision can nevertheless play a highly effective role in speeding up labelling efforts when initially constructing supervision information. We provide all data and full source code alongside an analysis and evaluation of the system.}
\end{abstract}
\vspace{-11pt}
\section{Introduction and Background}
\tbf{Holstein-Friesians are, with a global population of 70~million~\cite{faocattle} animals, the most numerous and also highest milk-yielding~\cite{tadesse2003milk} cattle breed in the world. Cattle identification (ID) via tags~\cite{houston2001computerised, buick2004animal, shanahan2009framework} is mandatory~\cite{eu82097, united2018states}, yet transponders~\cite{klindtworth1999electronic}, branding~\cite{adcock2018branding,awad2016classical} and biometric ID~\cite{kuhl2013} via face~\cite{barbedo2019study}, muzzle~\cite{petersen1922identification,kumar2017automatic, kimura2004structural, tharwat2014cattle, awad2019bag, el2015bovines}, retina~\cite{allen2008evaluation}, rear~\cite{qiao2019individual}, or coat patterns~\cite{martinez2013video,hu2020cow,li2017automatic, andrew2016automatic} are also viable. The last can conveniently operate from a distance above and has recently been implemented via supervised deep learning~\cite{andrew2017visual,andrew2019visual}. However, research into reducing manual labelling efforts for creating and maintaining such ID systems is in its infancy~\cite{andrew2020visual,vidal2021perspectives}. Particularly, unsupervised learning for coat pattern identification of Holstein-Friesians has not been tried and public datasets~\cite{OpenCows2020dataset} are small to date.}

\begin{figure}[t]
    \begin{center}
        \includegraphics[width=1\linewidth,height=190pt]{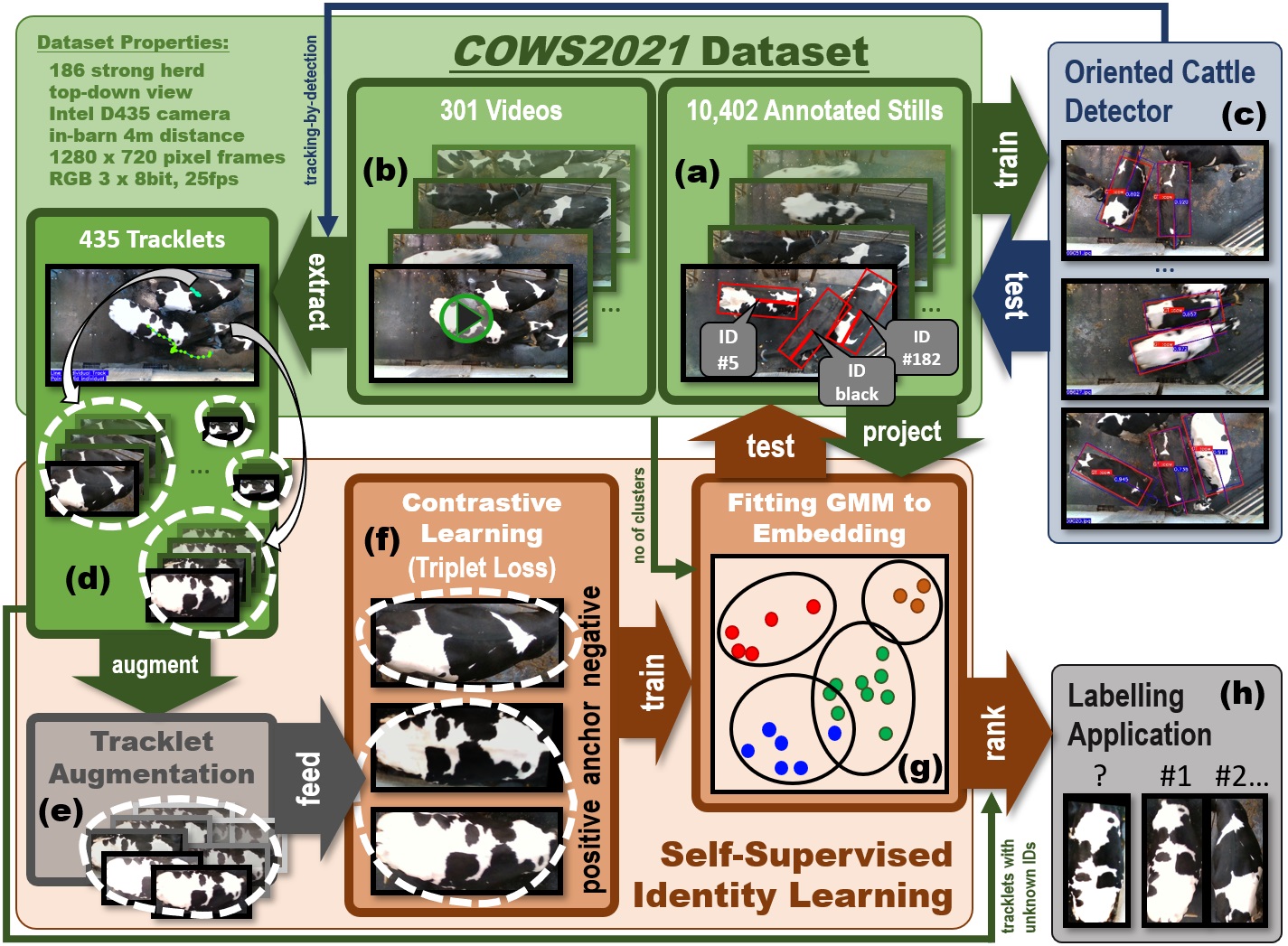}
    \end{center}
    \vspace{-16pt}
    \caption{\tbf{\small{$\bf{Conceptual}$ $\bf{Overview.}$ Our dataset $Cows2021$ provides both $(a)$~test images with oriented bounding-box and ID annotations, and $(b)$ unlabelled training videos of the same herd. $(c)$~ID-agnostic cattle tracking-by-detection across such videos yields $(d)$~scale and orientation-normalised tracklets, which are $(e)$~enhanced by augmentation. $(f)$~Frame-triplets with in-tracklet anchor and positive ROI vs. out-of-video negative ROI are used for contrastive learning of a latent embedding, wherein $(g)$ a GMM is fitted yielding an identity classifier by interpreting clusters as IDs. $(h)$~ID labelling applications for building productions systems from video datasets can significantly benefit from having a confidence-ranked list of possible identities provided to the user.}}}\vspace{-11pt}
\label{fig:pipeline}
\end{figure}
\tbf{This paper addresses these shortcomings and introduces the largest ID-annotated dataset of Holstein-Friesians: \textsl{Cows2021} so far, alongside a basic self-supervision system for video identification of individual animals~(see Fig.~\ref{fig:pipeline}).}

\begin{figure*}
    \begin{center}
    \includegraphics[width=\textwidth]{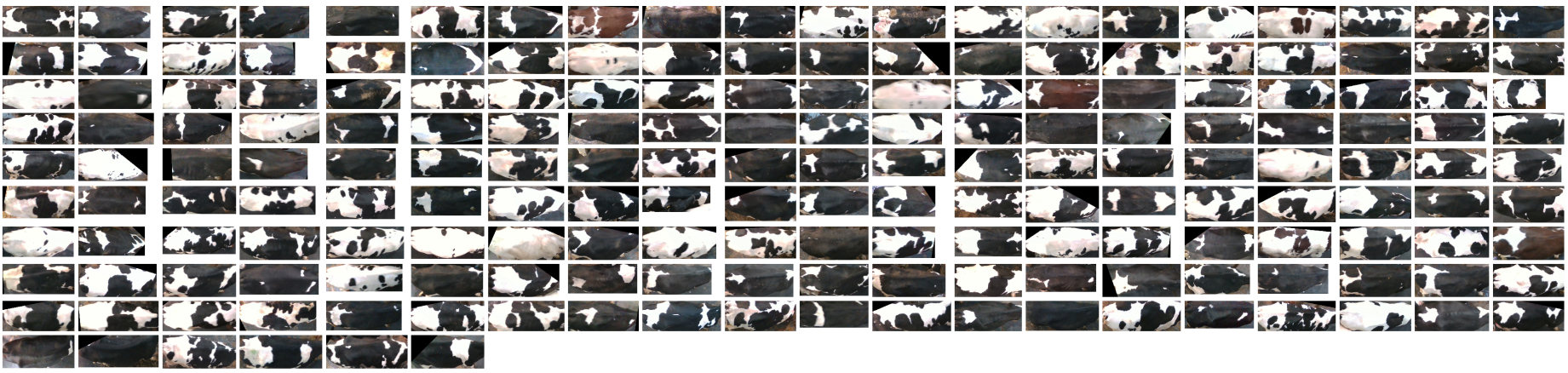}
    \end{center} 
    \vspace{-18pt}
    \caption{\tbf{\small{$\bf{The}$ $\textsl{Cows2021}$ $\bf{Herd.}$ Top-down, right facing view of the 186~individuals in the dataset normalised from RGB oriented bounding-box detections. Individually-characteristic black and white coat pattern patches are resolved at around $500\times 200$ pixels. 
    Note that $4$ animals ($2.2\%$ of herd) carry no white markings and are excluded as `un-enrollable' from the identification study.
    }}}\vspace{-8pt}
\label{fig:data_id}
\end{figure*}

\section{Dataset \textsl{Cows2021}}\label{sec:data}
\begin{figure}[b]
    \begin{center}
        \subfloat{\includegraphics[width=0.116\textwidth]{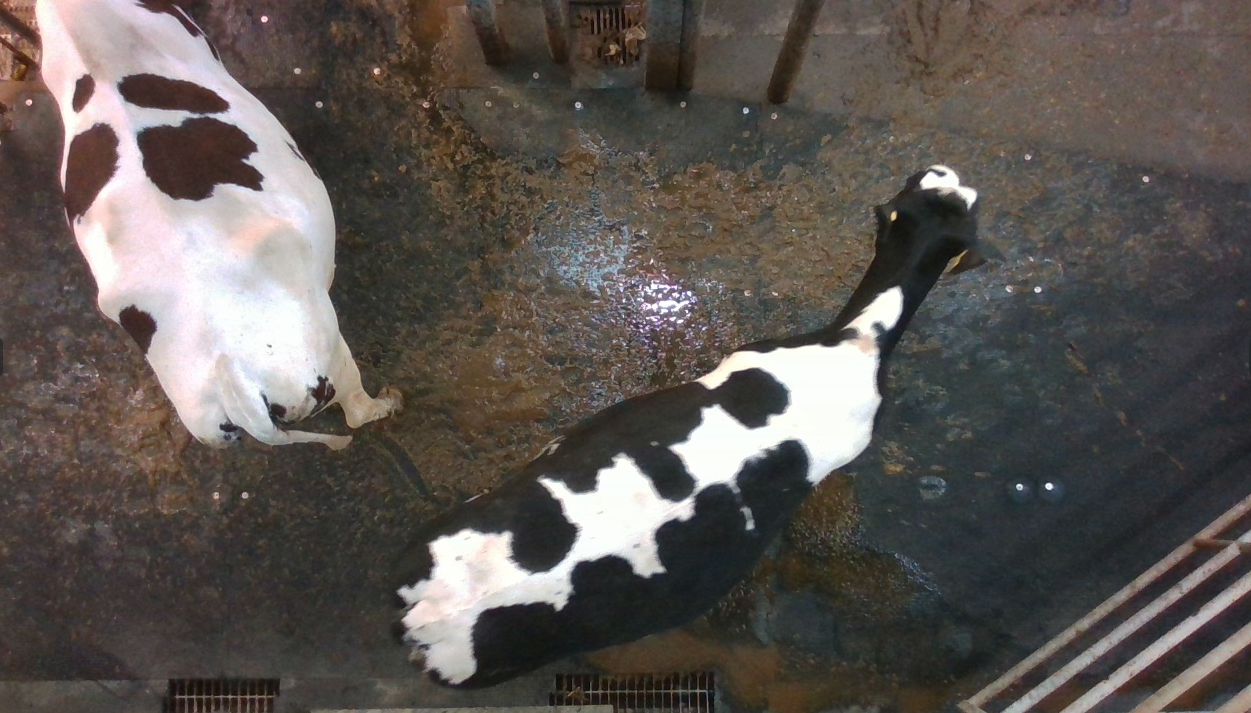}}
        \hfill
        \subfloat{\includegraphics[width=0.116\textwidth]{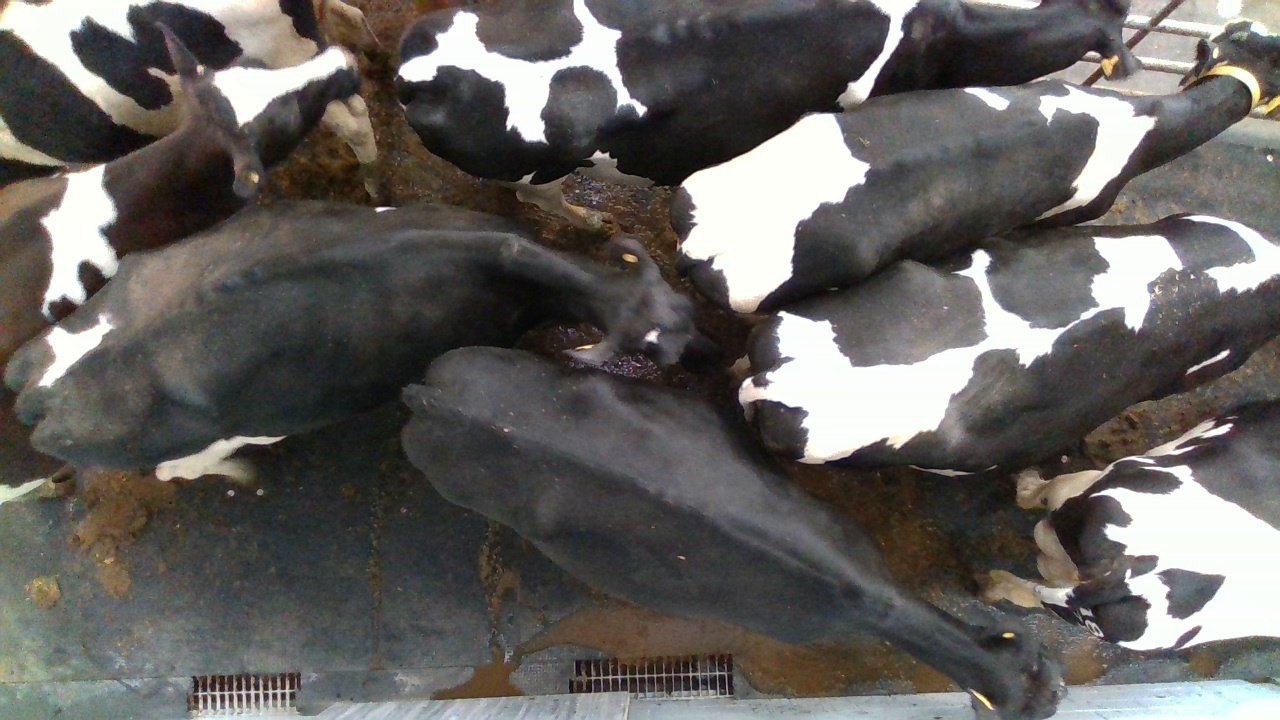}}
        \hfill
        \subfloat{\includegraphics[width=0.116\textwidth]{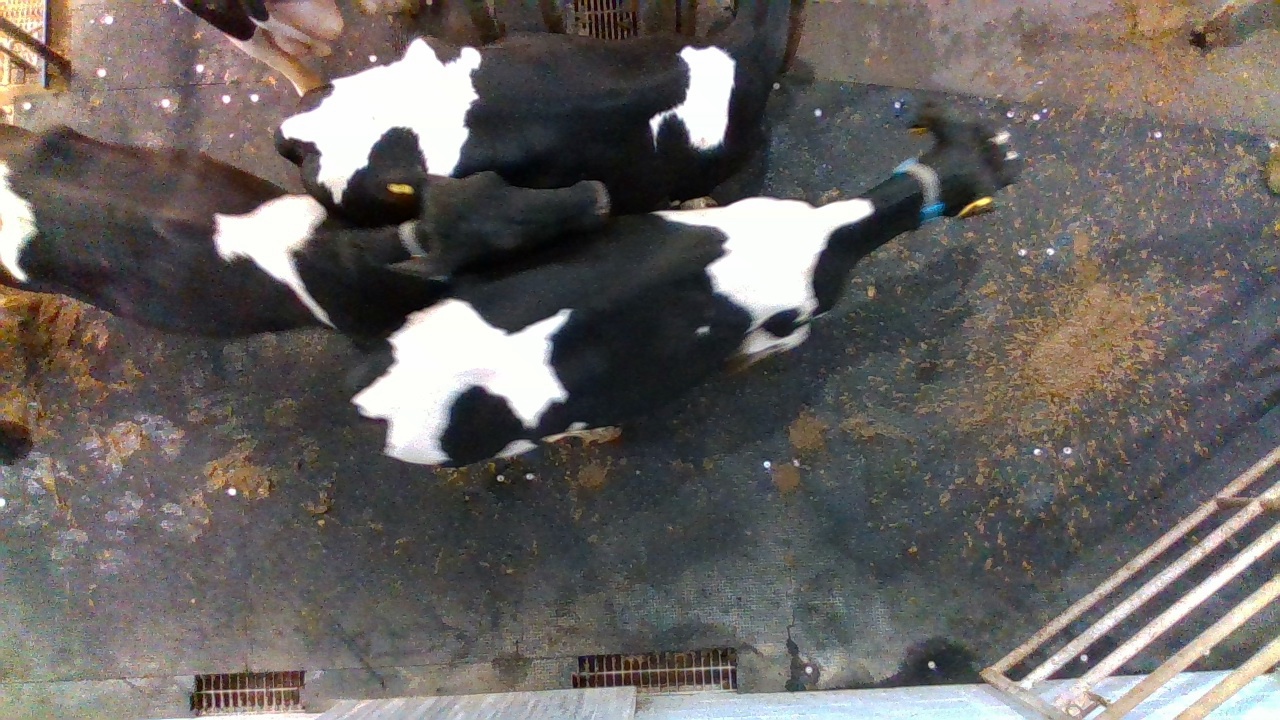}}
        \hfill
        \subfloat{\includegraphics[width=0.116\textwidth]{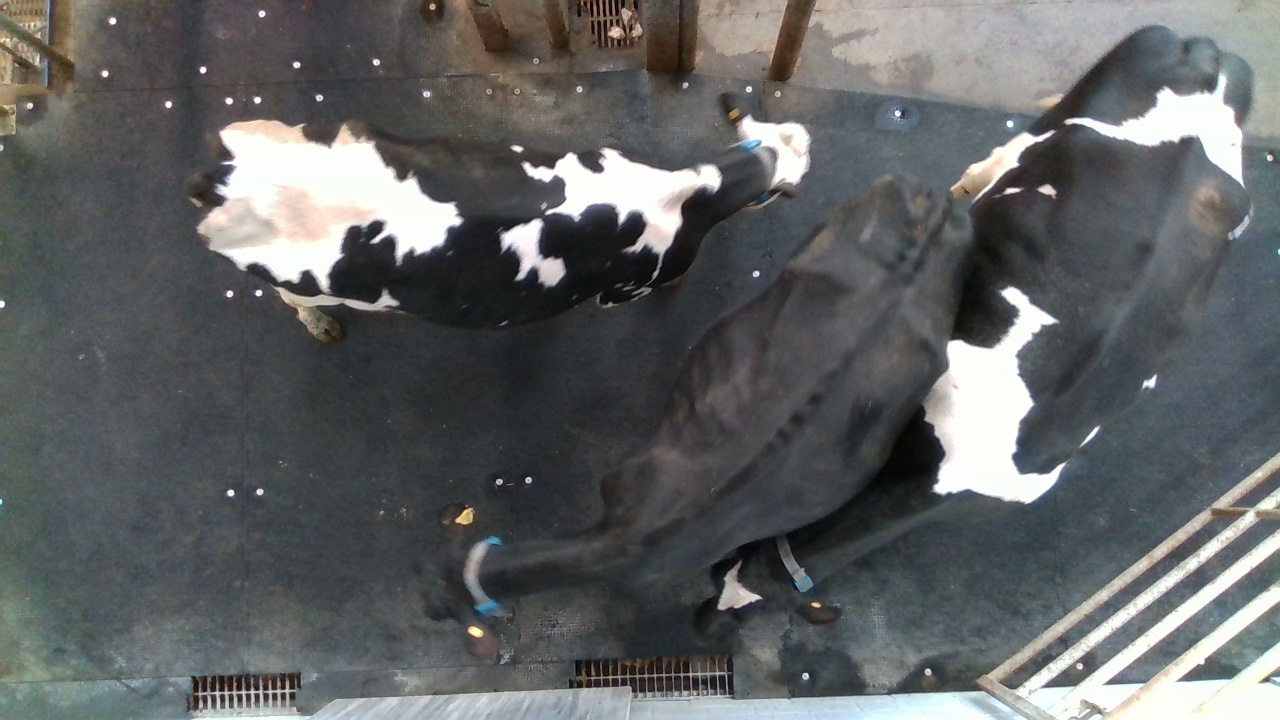}}
        \hfill
        \subfloat{\includegraphics[width=0.116\textwidth]{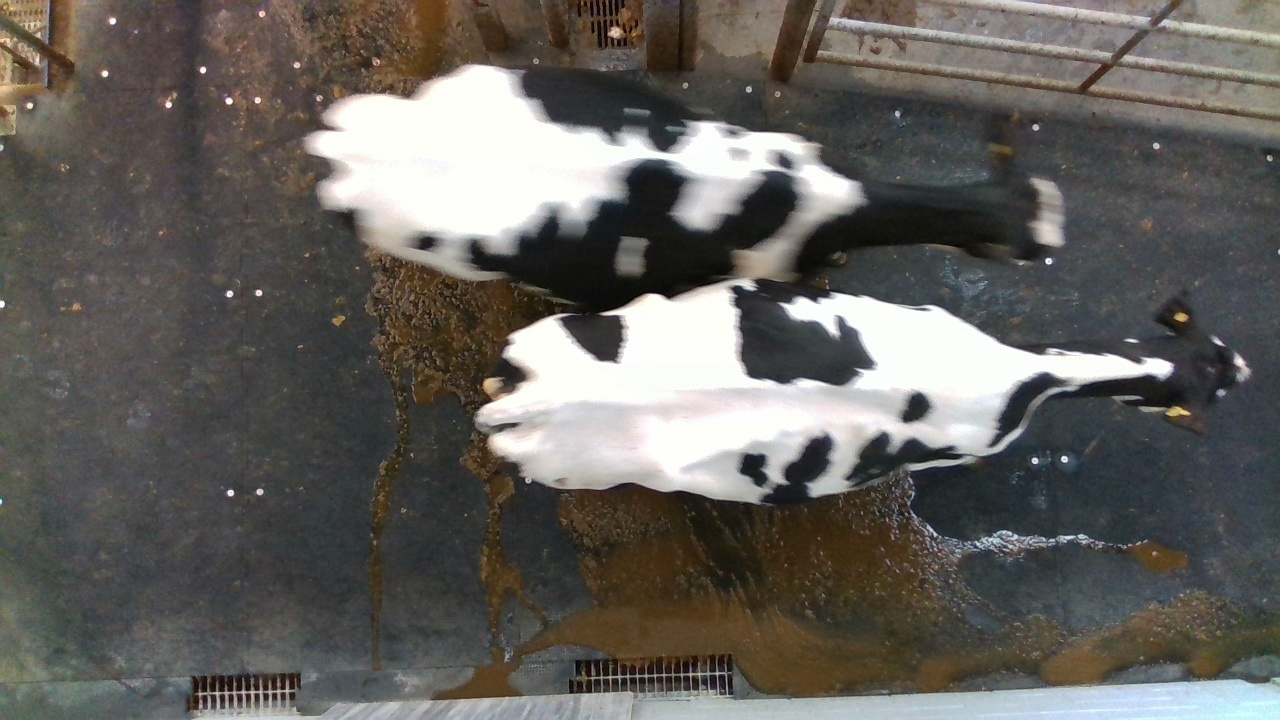}}
        \hfill
        \subfloat{\includegraphics[width=0.116\textwidth]{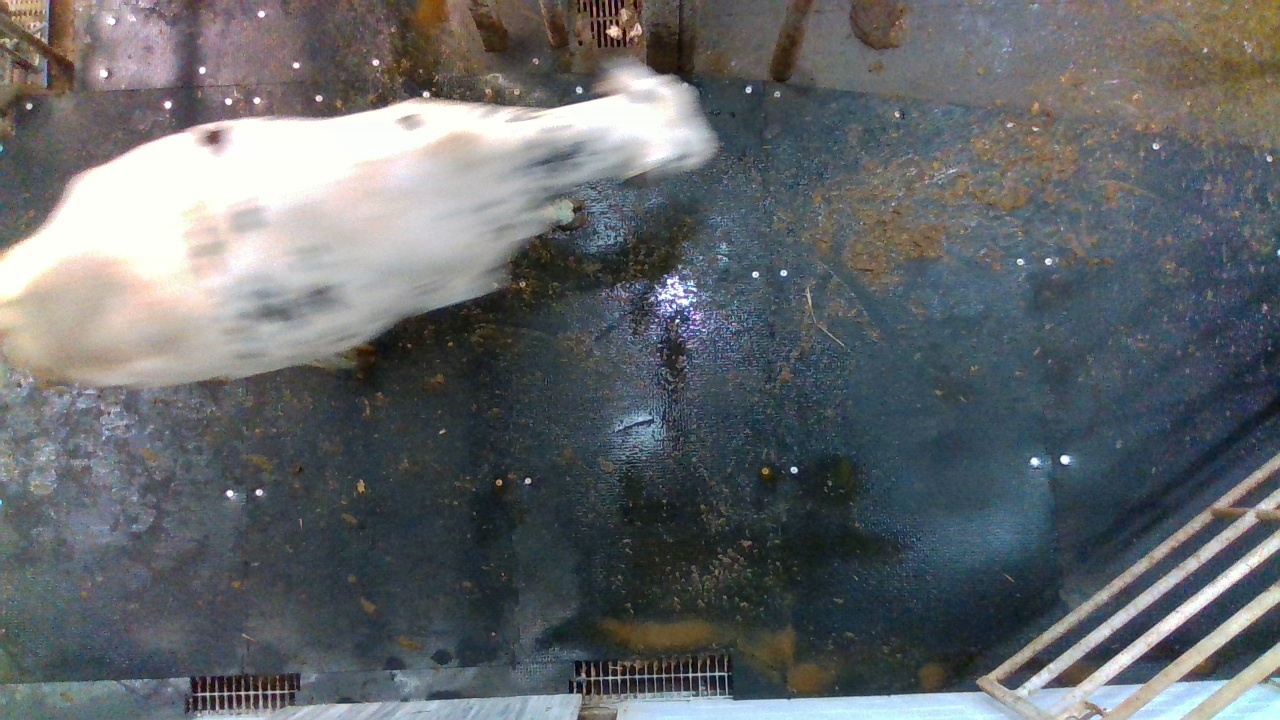}}
        \hfill
        \subfloat{\includegraphics[width=0.116\textwidth]{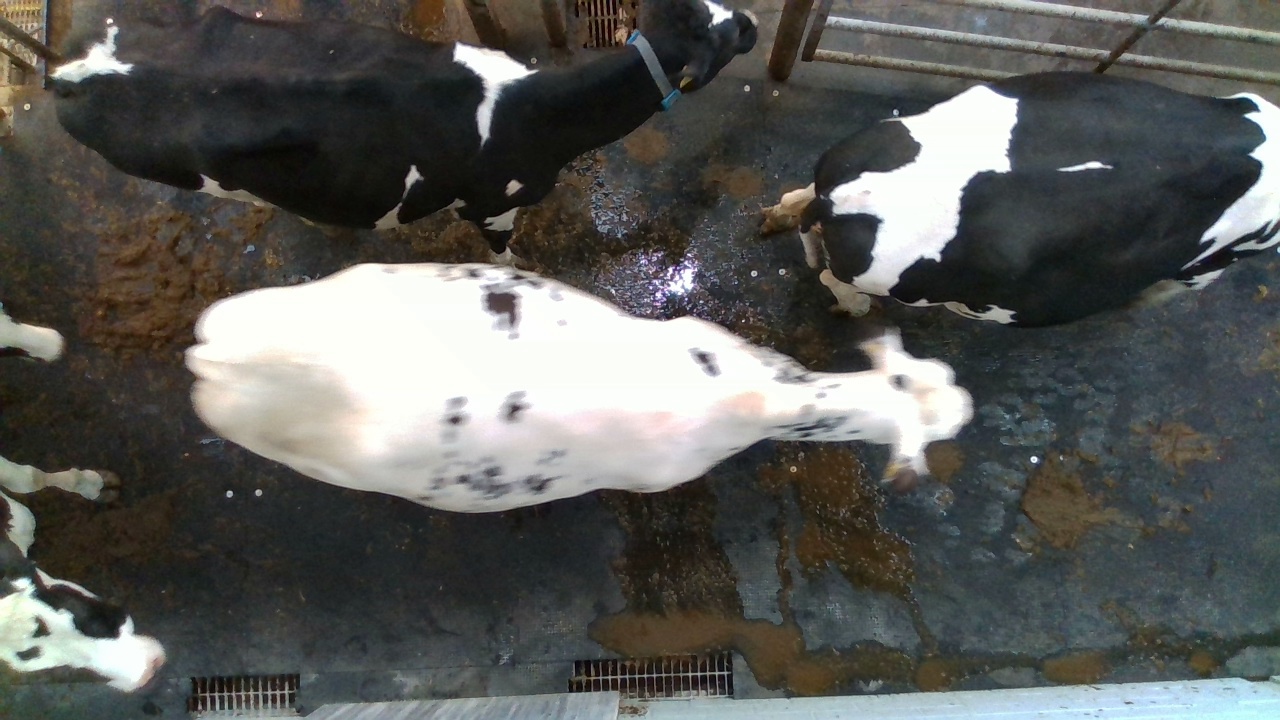}}
        \hfill
        \subfloat{\includegraphics[width=0.116\textwidth]{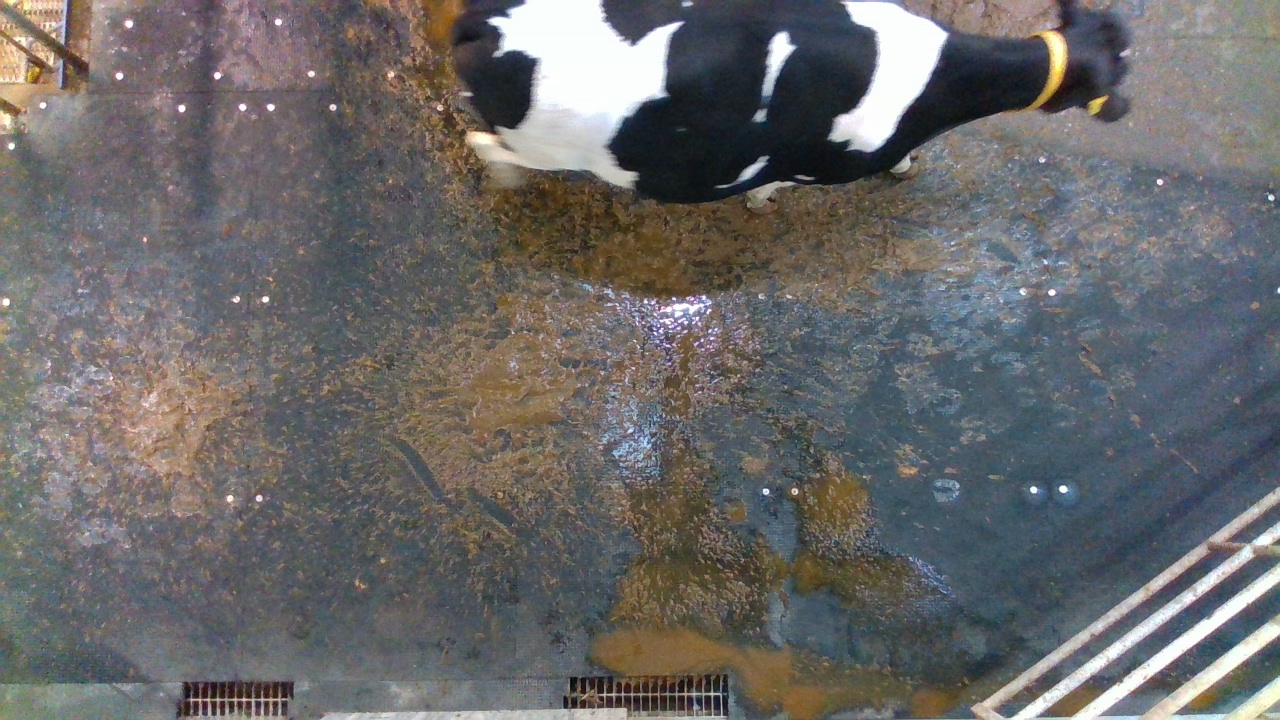}}
        \hfill
     \subfloat{\includegraphics[width=0.116\textwidth]{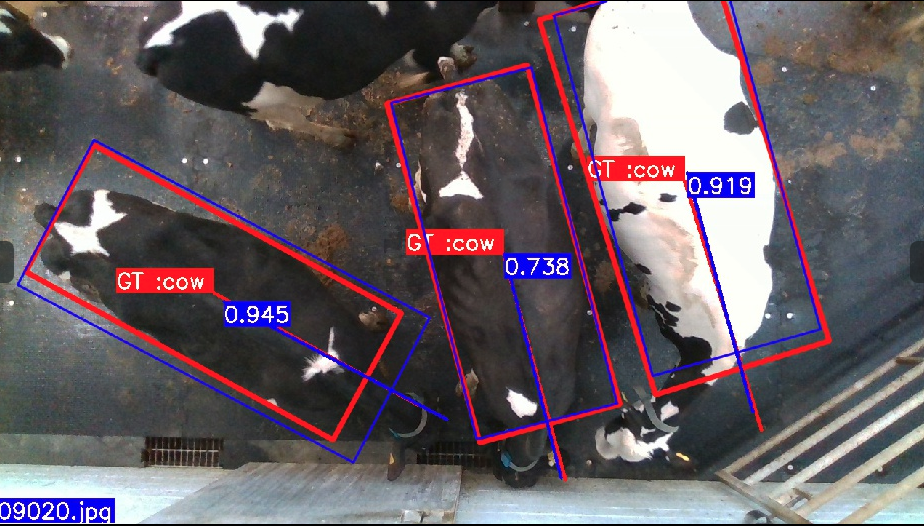}}
    \hfill
    \subfloat{\includegraphics[width=0.116\textwidth]{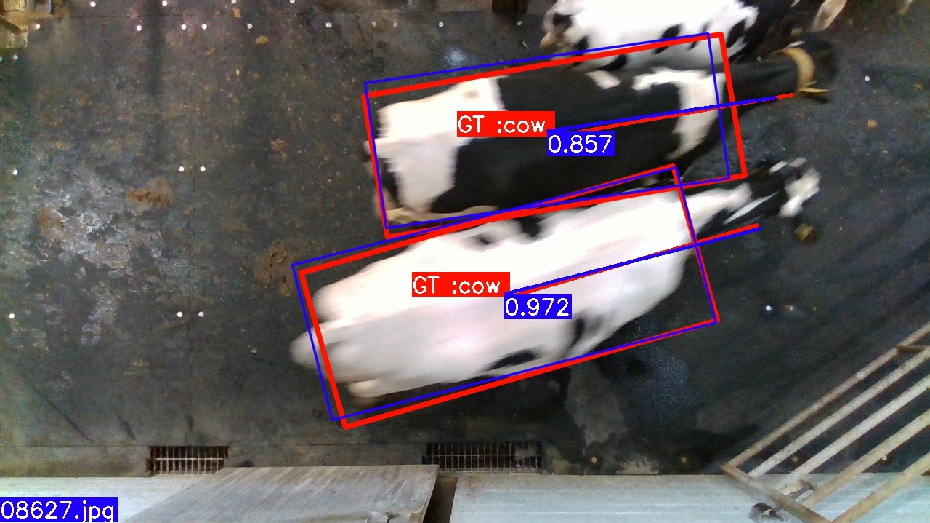}}
    \hfill
    \subfloat{\includegraphics[width=0.116\textwidth]{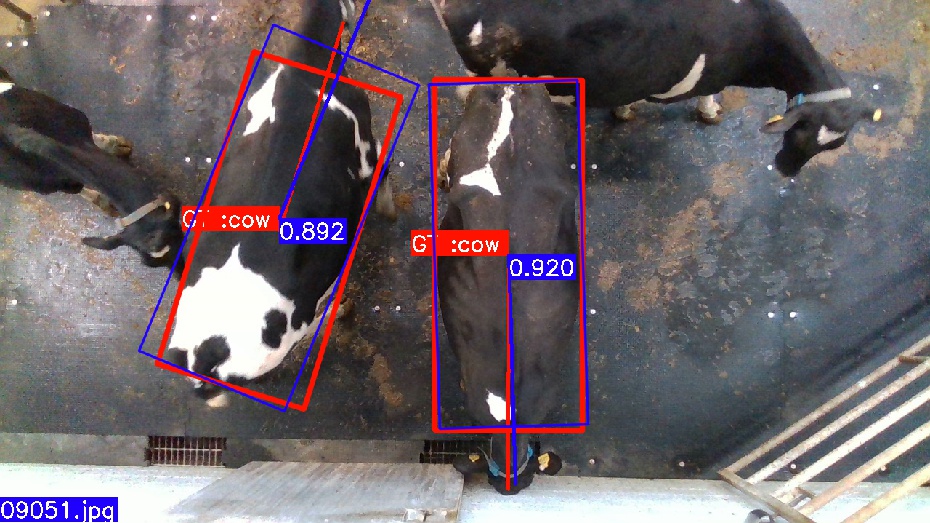}}
    \hfill
    \subfloat{\includegraphics[width=0.116\textwidth]{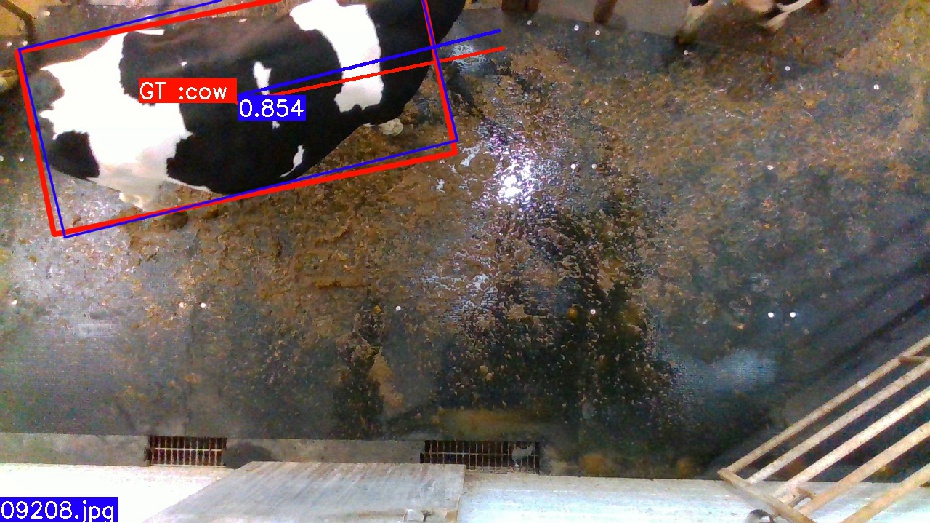}}
    \end{center}
    \vspace{-13pt}
    \caption{\tbf{\small{$\bf{Dataset}$ $\bf{and}$ $\bf{Cattle}$ $\bf{Localisation}$.} Representative frames characterising the datatset. $(top)$ Frames with varying animal orientation, crowding, clipping, and differing walking directions; $(middle)$ Frames with some motion blur, a mainly white cow with and without motion blur, and a near-boundary animal; $(bottom)$ Test images with oriented bounding-box annotations in red, output of ID-agnostic cattle detector~in~blue.}}\vspace{-11pt}
\label{fig:example_detection_dataset}
\end{figure}

\tbf{We introduce the RGB image dataset \textsl{Cows2021}\footnote{Available online at \url{https://data.bris.ac.uk}}, which features a herd of $186$~Holstein-Friesian cattle (see Fig.~\ref{fig:data_id}) and was acquired via an Intel D435 at University of Bristol’s Wyndhurst Farm in Langford Village, UK. The camera pointed downwards from $4m$ above the ground over a walkway~(see Fig.~\ref{fig:example_detection_dataset}) between milking parlour and holding pens. Motion-triggered recordings took place after milking across $1$~month of filming.}

\begin{figure}[t]
    \begin{center}
    \includegraphics[width=0.46\textwidth,height=100pt]{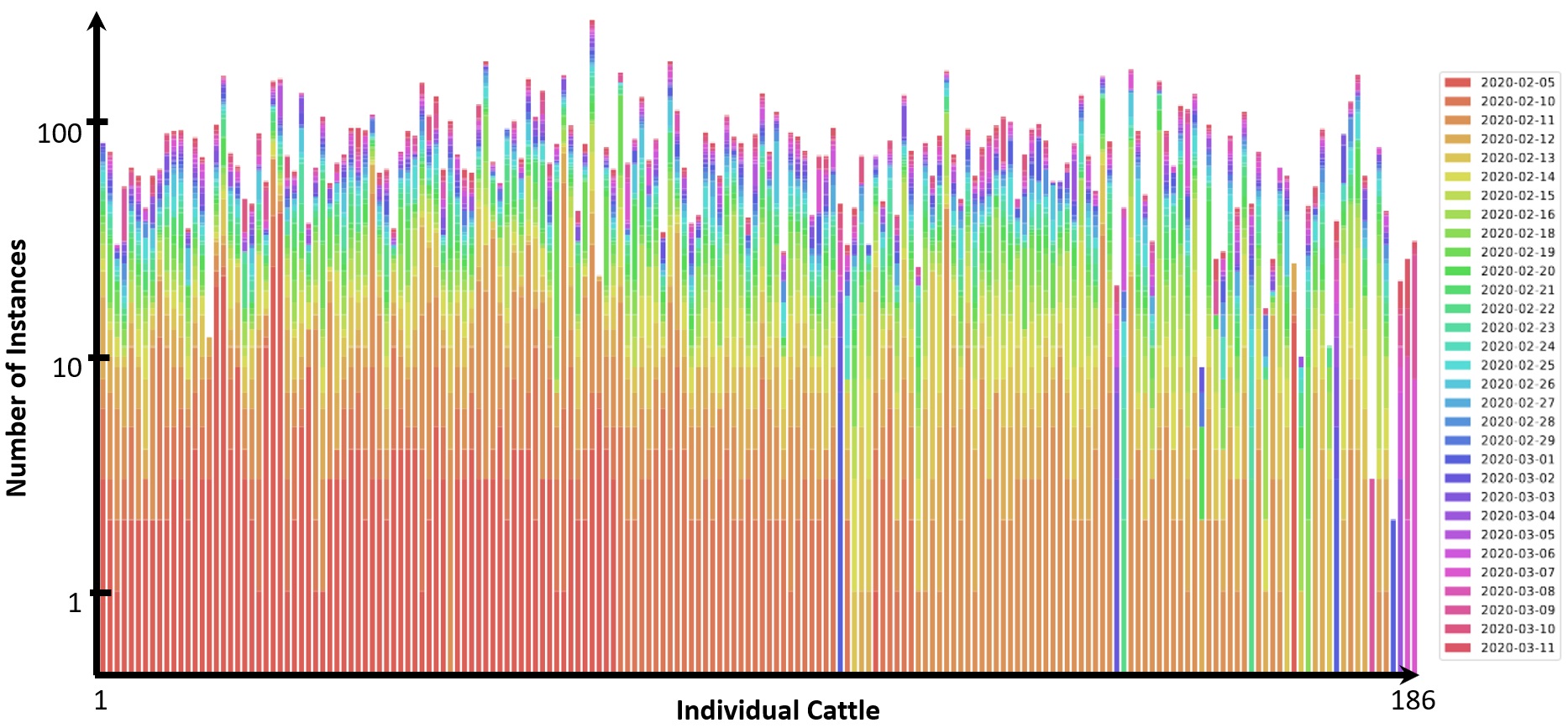}
    \end{center}
    \vspace{-15pt}
    \caption{\tbf{\small{$\bf{Acquisition}$ $\bf{Across}$ $\bf{Individuals.}$ Number of still images captured of the~$186$ individuals, with time of acquisition across the month of recording shown as colour values.}}}\vspace{-11pt}
\label{fig:data_graph}
\end{figure}

\tbf{The dataset is resolved at $1280\times 720$ pixels per frame with 8bit per RGB channel. It contains $10,402$ still images, in addition to $301$ videos (each of length  $~5.5s$) at 30fps. The distribution of stills across individuals and time reflects the natural workings of the farm~(see Fig.~\ref{fig:data_graph}). Various expert ground truth (GT) annotations are provided alongside the acquired dataset.}

{\bf Oriented Bounding-Box Cattle Annotations.}
\tbf{Adhering to the VOC 2012 guidelines~\cite{pascal-voc-2012} for object annotation, we manually labelled\footnote{Tool used: \url{https://github.com/cgvict/roLabelImg}} all visible cattle torso instances across the still image set. Annotations excluded the head, neck, legs and tail. Significantly clipped torso instances were (following ~\cite{pascal-voc-2012}) not used further and given a `clipped' tag. Example images from the resulting set of $13,165$ non-clipped cattle torso annotations are given in red in Fig.~\ref{fig:example_detection_dataset}~(bottom). Each oriented bounding-box label is parameterised by a tuple: $(c_x,c_y,w,h,\theta)$ corresponding to the box centrepoint, width, height, and head direction. }

{\bf Animal Identity Annotations.} 
\tbf{Overall~$13,784$ detected~(see Sec.~\ref{sec:detection}) cattle instances were manually ID-assigned to one of $182$ individuals (see Fig.~\ref{fig:data_id}). The $4$ all-black cows were excluded from the ID study subject to future research. The number of occurrences of individuals varies from~$2$ to~$273$ with mean~$\mu=77.5$ and standard deviation~$\sigma=39.9$~(see~Fig.~\ref{fig:data_graph}).} $8,670$ of these annotations were filmed on different days to the video data. These were used to form the identity test data. 

{\bf Video Data and Tracklet Annotations.} 
\tbf{In addition to still images, the dataset contains videos with tracklet information designed for utilisation as a rich source of self-supervision in identity learning.
Using a highly reliable ID-agnostic cattle detector~(see Sec.~\ref{sec:detection}) and sampling at 5Hz, tracking-by-detection was employed to connect nearest centrepoints of detections in neighbouring frames and thereby extract entire tracklets of the \textsl{same} individual~(see Fig.~\ref{fig:pipeline}). Manual checking ensured no tracking errors occurred. The average number of tracklets per video is $1.45$.}


\section{ID-agnostic Cattle Detector}\label{sec:detection}
\begin{figure}[t]
    \begin{center}
    \subfloat{\includegraphics[width=0.28\textwidth,height=110pt,trim=10 12 10 12, clip]{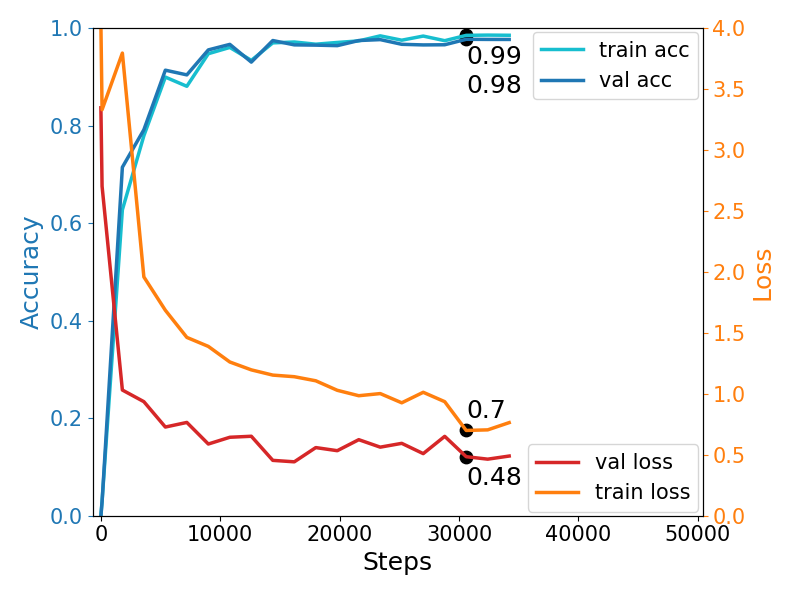}} 
    \subfloat{
    \footnotesize{
    \begin{tabular}{|c|c|}
        \hline
        \bf{Test AP} & \bf{{0.97}}  \\ \hline
        @IoU & 0.7  \\ 
        @Conf-Thr & 0.3 \\
        @NMS-Thr & 0.28 \\ \hline
        \#Train Images & 7,248 \\
        \#Val Images & 1,023 \\
        \#Test Images & 2,131 \\
        \hline 
    \end{tabular}  \vspace{12pt}
    } }
    \end{center}
    \vspace{-16pt}
    \caption{\tbf{\small{$\bf{Cattle}$ $\bf{Detector}$ $\bf{Performance.}$ Training and validation curves, working point (approx. @$29k$ steps), test Average Precision~(AP), and setup parameters for ID-agnostic cattle detector performing single frame oriented bounding-box detection.}}}\vspace{-11pt}
\label{fig:detector}
\end{figure}

\tbf{Existing multi-object single-frame cattle detectors~\cite{barbedo2019study, andrew2019aerial,andrew2020visual} produce image-aligned bounding-boxes that cannot avoid capturing several individuals in crowded scenes~(see~Fig.~\ref{fig:example_detection_dataset}), which is problematic for subsequent identity assignment. In response, we constructed a first orientation-aware cattle detector~(see Fig.~\ref{fig:example_detection_dataset} blue) by modifying RetinaNet~\cite{lin2017focal} with an ImageNet-pretrained~\cite{deng2009imagenet} ResNet50 backbone~\cite{he2016deep}. We added additional target parameters for orientation encoding and rotated anchors implemented in 5~layers~(P3~-~P7). To train the network, we partitioned the still image set approximately~$7:1:2$ for training, validation and testing, respectively. We used timestamps to split data so any temporal bias is reduced. We then trained the network against Focal Loss~\cite{lin2017focal} with settings $\gamma = 2$, $\alpha = 0.25$,  $\lambda = 1$ via SGD~\cite{robbins1951stochastic} with a learning rate of~$1 \times 10^{-5}$, momentum of~$0.9$~\cite{qian1999momentum}, and weight decay of~$1 \times 10^{-4}$. Fig.~\ref{fig:detector} illustrates training and depicts full performance benchmarks for the detector. For the test set, it operates at an Average Precision of $97.3\%$ using an Intersection over Union (IoU) threshold of $0.7$, reliably translating in-barn videos to tracklets.}

\section{Self-Supervised Animal Identity Learning}
\tbf{Given an ID-agnostic cattle detector~(see Sec.~\ref{sec:detection}), reliable tracklets can be generated~(see Sec.~\ref{sec:data}) from readily available in-barn videos of a Holstein-Friesian herd. We investigated how far this data can be used to self-supervise the learning of filmed individual animals to aid the time-consuming task of manual labelling.}
\subsection{Contrastive Training} 
\begin{figure}[t]
    \begin{center}
    \subfloat{\includegraphics[width=0.34\textwidth,height=110pt,trim=10 12 10 12, clip]{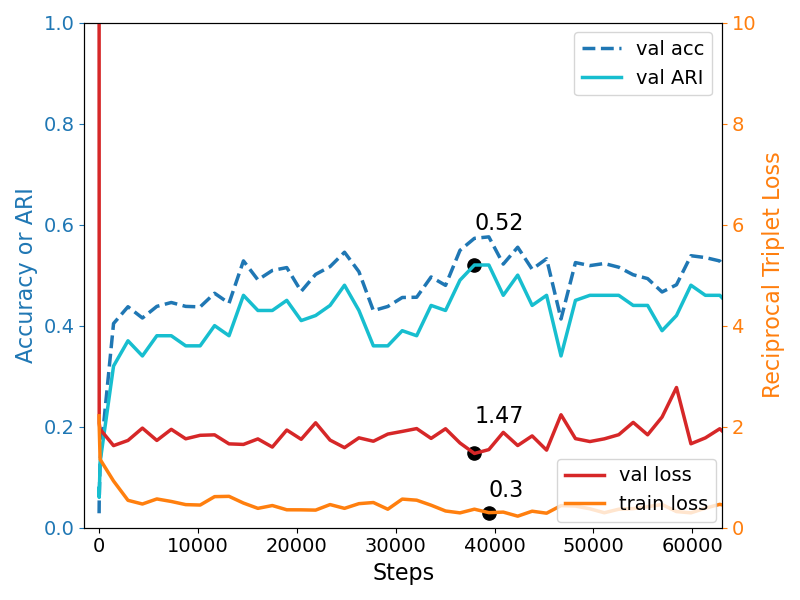}}   
    \end{center}
    \vspace{-16pt}
    \caption{\tbf{\small{$\bf{Self}$-$\bf{Supervised}$ $\bf{Identity}$ $\bf{Learning.}$ Training and validation curves, working point (approx. @38k Steps), accuracy and Adjusted Rand Index benchmarks for learning cattle identities via triplet loss.}}}\vspace{-11pt}
\label{fig:identity}
\end{figure}

\tbf{{\bf Identification Network and Triplet Loss.} We use a ResNet50~\cite{he2016deep} pretrained on ImageNet~\cite{deng2009imagenet}, modified to have a fully-connected final layer to learn a latent $128$-dimensional ID-space. Across the training data of all videos, we normalise each tracklet for rotation~(as seen in Fig.~\ref{fig:data_id}) and organise it into a `positive' ID sample set representing the same, unknown individual. We pair this set against `negative' samples from random cattle of other videos, which have a high chance of containing a different individual. All sets are enhanced via rotational augmentation~(max.\ angle $\pm{7}^{\circ}$). The separate image data was used as a validation and testing base, split $1 : 3 $. Reciprocal triplet loss~(RTL)~\cite{masullo2019goes} is then employed for learning an ID-encoding latent space via an online batch hard mining strategy~\cite{hermans2017defense}:}\vspace{-6pt}
\begin{equation}
\label{eq:reciprocal-triplet-loss}
    \mathbb{L}_{RTL} = d(x_a, x_p) + \frac{1}{d(x_a, x_n)}\vspace{-2pt}
\end{equation}
\tbf{where~$x_a$ and~$x_p$ are sampled from the `positive' set and~$x_n$ is a `negative' sample.  We trained the network for 7 hours via SGD~\cite{robbins1951stochastic} over $50$ epochs with batch size~$16$, learning rate~$1 \times 10^{-3}$, margin $\alpha=2$, and weight decay~$1 \times 10^{-4}$. The pocket algorithm~\cite{stephen1990perceptron} against the validation set was used to tackle overfitting~(see Fig.~\ref{fig:identity}).} 

\subsection{Animal Identity Discovery via Clustering} 
\begin{figure}[t]    \vspace{-23pt}
    \begin{center}
        \subfloat{\includegraphics[width=0.38\textwidth,height=170pt,trim=35 35 35 35, clip]{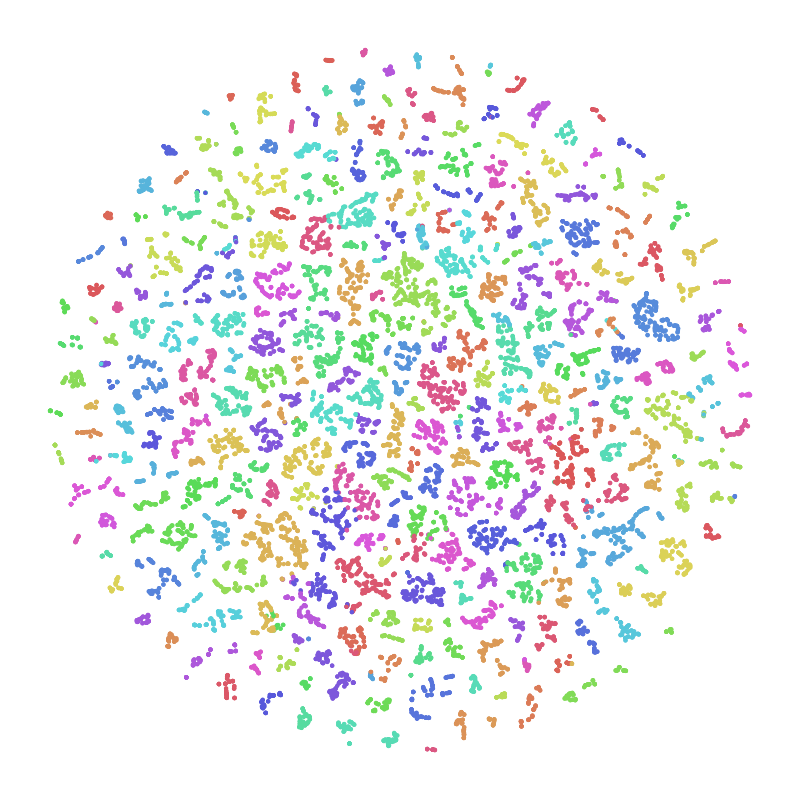}} 
    \end{center}
    \vspace{-13pt}
    \caption{\tbf{\small{$\bf{Training}$ $\bf{Embeddings.}$ t-SNE plot of training data projected into the latent space and partitioned by the GMM into~$182$ identity clusters shown using random colours}.}}\vspace{-11pt}
\label{fig:training_embeddings}
\end{figure}

\tbf{{\bf Clustering.} We then fitted~{\cite{scikit-learn} a Gaussian Mixture Model~(GMM)~\cite{reynolds2009gaussian} }to the generated $128$-dimensional space by setting the cluster cardinality to the known~$k=182$ patterned individual animals with $200$ iterations. Resulting clusters are then interpreted as representing separate animal identities. A t-distributed Stochastic Neighbour Embedding~(t-SNE)~\cite{van2008visualizing} of the training set projected into the clustered space is visualised in Fig.~\ref{fig:training_embeddings}. In order to evaluate the clustering performance, we used two measures: the Adjusted Rand Index~(ARI)~\cite{hubert1985comparing} and ID prediction accuracy. For the latter, each GMM cluster is assigned to the one individual ID with the highest overlap which is defined as:}  \vspace{-6pt}
\begin{equation}
\label{eq:overlap}
\mathbb{O}_{l}=C / L\vspace{-2pt}
\end{equation}
\tbf{where $C$ is the number of images in a GMM cluster that belong to an individual, and $L$ is the total number of images of the individual. This produces (GMM Cluster)-(ID Label) pairs for accuracy evaluation.}



\begin{figure}[b]
    \centering
    \vspace{-11pt}
    \includegraphics[width=0.43\textwidth,height=150pt,trim= 35 32 65 30, clip]{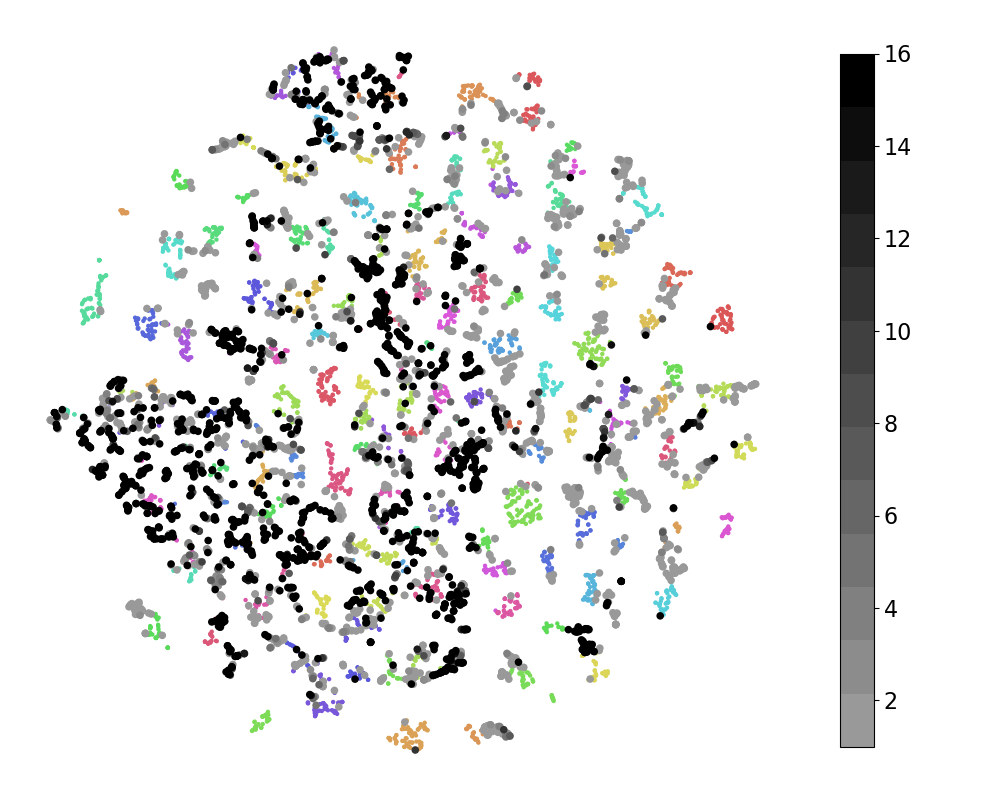} 
    \caption{\small{\tbf{$\bf{Clustered}$ $\bf{Embedding.}$ t-SNE plot across test images. Colour indicates correct ID assignments of test data points, gray-to-black indicates Top-N severity of mismatch.}}}\vspace{-11pt}
    \label{fig:tsne-class-overlay}
\end{figure}

\tbf{{\bf Top-N Accuracy.} In order to quantitatively evaluate the capacity to aid human annotation, we consider a scenario where a user annotates IDs as a one-out-of-N pick (expanding N if the correct ID is not present). Thus, the Top-N system accuracy~\cite{NIPS2012_4824} is a key measure to investigate. For each cluster one can rank all identities according to~$\mathbb{O}_{l}$. Identities that have a~$\mathbb{O}_{l}=0$ form the randomly assigned tail of the sequence. For every data point this provides a general Top-N assigned ID. Finding the GT identity amongst the Top-N assigned IDs is then counted as correct identification.}
    
\section{Experimental Results and Discussion}
\tbf{{\bf Structural Clustering Similarity.} In order to characterise the ID performance as if this were a new, unknown herd, we calculated the ARI to be $0.53$ for the test set when measured between the partitioning derived from the clustering provided by the GMM versus the identity GT. This measure captures the (purely structural) similarities between the two clusterings.}

\tbf{{\bf Clustering Accuracy.} In order to characterise the ID performance with class labels, we calculated Top-N accuracy for the test set as depicted in~Table~\ref{table}. Figure~\ref{fig:tsne-class-overlay} visualises the identification performance and misclassification severity using a t-SNE plot.
}

\tbf{{\bf Context and Result Discussion.} Considering that $182$ classes were used and absolutely no training labelling was provided, results of~$57.0\%$ {Top-1 accuracy and~$76.9\%$ T}op-4 accuracy are an encouraging and practically relevant first step towards self-supervision in this domain. We know that individual Holstein-Friesian identification via \emph{supervised} deep learning is a widely solved task with systems achieving near-perfect benchmarks when using multi-frame LRCNs~\cite{andrew2019aerial} and good results even in partial annotation settings~\cite{andrew2020visual}. However, labelling efforts are laborious for supervised systems of larger herds; they require days if not weeks of manual annotation effort using visual dictionaries of animal ground truth. Humans can efficiently compare small sets of images. Thus, using the described pipeline we could present the user with a set of (e.g.\ $4$) images that contain the correct individual with a chance better than 3-in-4. As part of a toolchain, the approach presented can potentially dramatically reduce labelling times and help bootstrap production systems via combinations of self-supervised learning followed by open set fine-tuning~\cite{andrew2020visual}.
}

\begin{table}[hb]
\begin{center}
    \begin{tabular}{|l|c|c|c|c|c|}
        \hline
        Top-N & N=1 & N=2 & N=4 & N=8  & N=16  \\ \hline\hline
        Accuracy~(\%) & 57.0 & 71.8 & 76.9 & 79.7 & 81.8  \\ \hline
        
    \end{tabular}
\end{center}\vspace{-16pt}
    \caption{\small{\tbf{$\bf{Top-N}$ $\bf{Performance.}$ Shown is ID accuracy for a variety of $N$ across all $8,670$ test instances of the $182$ identities. }}}\vspace{-11pt}
\label{table}
\end{table}

\section{Conclusion}{

\tbf{In this paper we presented the largest identity-annotated Holstein-Friesian cattle dataset, \textsl{Cows2021}, made available to date. We also showed a first self-supervision framework for identifying individual animals. Driven by the enormous labelling effort involved in constructing visual cattle identification systems, we proposed exploiting coat pattern appearance across videos as a self-supervision signal. A generic cattle detector yielded oriented bounding-boxes which were normalised and augmented. Triplet loss contrastive learning was then used to construct a latent space wherein we fitted a GMM. This yielded a cattle identity classifier which we evaluated. \ Our results showed that the achieved accuracy levels are strong enough to help speed up ID labelling efforts for supervised systems in the future. Despite the need for even larger datatsets, we hope that the published dataset, code, and benchmark will stimulate research in the area of self-supervision learning for biometric animal (re)identification.}}

\footnotesize
{~\\ \bf Acknowledgements.} This work was supported by The Alan Turing Institute under the EPSRC grant EP/N510129/1 and the John Oldacre Foundation through the John Oldacre Centre for Sustainability and Welfare in Dairy Production, Bristol Veterinary School. Jing Gao was supported by the China Scholarship Council. We thank Kate Robinson and the Wyndhurst Farm staff for their assistance with data collection.

\clearpage
{\footnotesize
\bibliographystyle{ieee_fullname}
\bibliography{egbib}

\begin{thebibliography}{10}\itemsep=-1pt

\bibitem{adcock2018branding}
Sarah~JJ Adcock, Cassandra~B Tucker, Gayani Weerasinghe, and Eranda Rajapaksha.
\newblock Branding practices on four dairies in kantale, sri lanka.
\newblock {\em Animals}, 8(8):137, 2018.

\bibitem{allen2008evaluation}
A Allen, B Golden, M Taylor, D Patterson, D Henriksen, and R Skuce.
\newblock Evaluation of retinal imaging technology for the biometric
  identification of bovine animals in northern ireland.
\newblock {\em Livestock science}, 116(1-3):42--52, 2008.

\bibitem{andrew2019visual}
William Andrew.
\newblock {\em Visual biometric processes for collective identification of
  individual Friesian cattle}.
\newblock PhD thesis, University of Bristol, 2019.

\bibitem{OpenCows2020dataset}
Will Andrew, Tilo Burghardt, Neill Campbell, and Jing Gao.
\newblock The opencows2020 dataset, 2020.
\newblock https://data.bris.ac.uk/data/dataset/10m32xl88x2b61zlkkgz3fml17.

\bibitem{andrew2020visual}
William Andrew, Jing Gao, Neill Campbell, Andrew~W Dowsey, and Tilo Burghardt.
\newblock Visual identification of individual holstein friesian cattle via deep
  metric learning.
\newblock {\em arXiv preprint arXiv:2006.09205}, 2020.

\bibitem{andrew2017visual}
William Andrew, Colin Greatwood, and Tilo Burghardt.
\newblock Visual localisation and individual identification of holstein
  friesian cattle via deep learning.
\newblock In {\em Proceedings of the IEEE International Conference on Computer
  Vision}, pages 2850--2859, 2017.

\bibitem{andrew2019aerial}
William Andrew, Colin Greatwood, and Tilo Burghardt.
\newblock Aerial animal biometrics: Individual friesian cattle recovery and
  visual identification via an autonomous uav with onboard deep inference.
\newblock In {\em 2019 IEEE/RSJ International Conference on Intelligent Robots
  and Systems (IROS)}, pages 237--243. IEEE, 2019.

\bibitem{andrew2016automatic}
William Andrew, Sion Hannuna, Neill Campbell, and Tilo Burghardt.
\newblock Automatic individual holstein friesian cattle identification via
  selective local coat pattern matching in rgb-d imagery.
\newblock In {\em 2016 IEEE International Conference on Image Processing
  (ICIP)}, pages 484--488. IEEE, 2016.

\bibitem{awad2016classical}
Ali~Ismail Awad.
\newblock From classical methods to animal biometrics: A review on cattle
  identification and tracking.
\newblock {\em Computers and Electronics in Agriculture}, 123:423--435, 2016.

\bibitem{awad2019bag}
Ali~Ismail Awad and M Hassaballah.
\newblock Bag-of-visual-words for cattle identification from muzzle print
  images.
\newblock {\em Applied Sciences}, 9(22):4914, 2019.

\bibitem{barbedo2019study}
Jayme Garcia~Arnal Barbedo, Luciano~Vieira Koenigkan, Thiago~Teixeira Santos,
  and Patr{\'\i}cia~Menezes Santos.
\newblock A study on the detection of cattle in uav images using deep learning.
\newblock {\em Sensors}, 19(24):5436, 2019.

\bibitem{buick2004animal}
W Buick.
\newblock Animal passports and identification.
\newblock {\em Defra Veterinary Journal}, 15:20--26, 2004.

\bibitem{deng2009imagenet}
Jia Deng, Wei Dong, Richard Socher, Li-Jia Li, Kai Li, and Li Fei-Fei.
\newblock Imagenet: A large-scale hierarchical image database.
\newblock In {\em 2009 IEEE conference on computer vision and pattern
  recognition}, pages 248--255. Ieee, 2009.

\bibitem{el2015bovines}
Hagar~M El~Hadad, Hamdi~A Mahmoud, and Farid~Ali Mousa.
\newblock Bovines muzzle classification based on machine learning techniques.
\newblock {\em Procedia Computer Science}, 65:864--871, 2015.

\bibitem{pascal-voc-2012}
M. Everingham, L. Van~Gool, C.~K.~I. Williams, J. Winn, and A. Zisserman.
\newblock The {PASCAL} {V}isual {O}bject {C}lasses {C}hallenge 2012 {(VOC2012)}
  {R}esults.
\newblock
  http://www.pascal-network.org/challenges/VOC/voc2012/workshop/index.html.

\bibitem{faocattle}
Food and Agriculture~Organization of~the United~Nations.
\newblock Gateway to dairy production and products.
\newblock
  \url{http://www.fao.org/dairy-production-products/production/dairy-animals/cattle/en/}.
\newblock [Online; accessed 4-August-2020].

\bibitem{he2016deep}
Kaiming He, Xiangyu Zhang, Shaoqing Ren, and Jian Sun.
\newblock Deep residual learning for image recognition.
\newblock In {\em Proceedings of the IEEE conference on computer vision and
  pattern recognition}, pages 770--778, 2016.

\bibitem{hermans2017defense}
Alexander Hermans, Lucas Beyer, and Bastian Leibe.
\newblock In defense of the triplet loss for person re-identification.
\newblock {\em arXiv preprint arXiv:1703.07737}, 2017.

\bibitem{houston2001computerised}
R Houston.
\newblock A computerised database system for bovine traceability.
\newblock {\em Revue Scientifique et Technique-Office International des
  Epizooties}, 20(2):652, 2001.

\bibitem{hu2020cow}
Hengqi Hu, Baisheng Dai, Weizheng Shen, Xiaoli Wei, Jian Sun, Runze Li, and
  Yonggen Zhang.
\newblock Cow identification based on fusion of deep parts features.
\newblock {\em Biosystems Engineering}, 192:245--256, 2020.

\bibitem{hubert1985comparing}
Lawrence Hubert and Phipps Arabie.
\newblock Comparing partitions.
\newblock {\em Journal of classification}, 2(1):193--218, 1985.

\bibitem{kimura2004structural}
Akio Kimura, Kazushi Itaya, and Takashi Watanabe.
\newblock Structural pattern recognition of biological textures with growing
  deformations: A case of cattle's muzzle patterns.
\newblock {\em Electronics and Communications in Japan (Part II: Electronics)},
  87(5):54--66, 2004.

\bibitem{klindtworth1999electronic}
M Klindtworth, G Wendl, K Klindtworth, and H Pirkelmann.
\newblock Electronic identification of cattle with injectable transponders.
\newblock {\em Computers and electronics in agriculture}, 24(1-2):65--79, 1999.

\bibitem{NIPS2012_4824}
Alex Krizhevsky, Ilya Sutskever, and Geoffrey~E. Hinton.
\newblock Imagenet classification with deep convolutional neural networks.
\newblock In F. Pereira, C.~J.~C. Burges, L. Bottou, and K.~Q. Weinberger,
  editors, {\em Advances in Neural Information Processing Systems 25}, pages
  1097--1105. Curran Associates, Inc., 2012.

\bibitem{kuhl2013}
Hjalmar~S K{\"u}hl and Tilo Burghardt.
\newblock Animal biometrics: quantifying and detecting phenotypic appearance.
\newblock {\em Trends in ecology \& evolution}, 28(7):432--441, 2013.

\bibitem{kumar2017automatic}
Santosh Kumar and Sanjay~Kumar Singh.
\newblock Automatic identification of cattle using muzzle point pattern: a
  hybrid feature extraction and classification paradigm.
\newblock {\em Multimedia Tools and Applications}, 76(24):26551--26580, 2017.

\bibitem{li2017automatic}
Wenyong Li, Zengtao Ji, Lin Wang, Chuanheng Sun, and Xinting Yang.
\newblock Automatic individual identification of holstein dairy cows using
  tailhead images.
\newblock {\em Computers and electronics in agriculture}, 142:622--631, 2017.

\bibitem{lin2017focal}
Tsung-Yi Lin, Priya Goyal, Ross Girshick, Kaiming He, and Piotr Doll{\'a}r.
\newblock Focal loss for dense object detection.
\newblock In {\em Proceedings of the IEEE international conference on computer
  vision}, pages 2980--2988, 2017.

\bibitem{martinez2013video}
Carlos~A Martinez-Ortiz, Richard~M Everson, and Toby Mottram.
\newblock Video tracking of dairy cows for assessing mobility scores.
\newblock 2013.

\bibitem{masullo2019goes}
Alessandro Masullo, Tilo Burghardt, Dima Damen, Toby Perrett, and Majid
  Mirmehdi.
\newblock Who goes there? exploiting silhouettes and wearable signals for
  subject identification in multi-person environments.
\newblock In {\em Proceedings of the IEEE International Conference on Computer
  Vision Workshops}, pages 0--0, 2019.

\bibitem{united2018states}
United States~Department of Agriculture (USDA)~Animal and Plant
  Health~Inspection Service.
\newblock Cattle identification.
\newblock
  \url{https://www.aphis.usda.gov/aphis/ourfocus/animalhealth/nvap/NVAP-Reference-Guide/Animal-Identification/Cattle-Identification}.
\newblock [Online; accessed 14-November-2018].

\bibitem{eu82097}
European Parliament and Council.
\newblock Establishing a system for the identification and registration of
  bovine animals and regarding the labelling of beef and beef products and
  repealing council regulation (ec) no 820/97.
\newblock
  \url{http://eur-lex.europa.eu/legal-content/EN/TXT/?uri=celex:32000R1760},
  1997.
\newblock [Online; accessed 29-January-2016].

\bibitem{scikit-learn}
F. Pedregosa, G. Varoquaux, A. Gramfort, V. Michel, B. Thirion, O. Grisel, M.
  Blondel, P. Prettenhofer, R. Weiss, V. Dubourg, J. Vanderplas, A. Passos, D.
  Cournapeau, M. Brucher, M. Perrot, and E. Duchesnay.
\newblock Scikit-learn: Machine learning in {P}ython.
\newblock {\em Journal of Machine Learning Research}, 12:2825--2830, 2011.

\bibitem{petersen1922identification}
WE Petersen.
\newblock The identification of the bovine by means of nose-prints.
\newblock {\em Journal of dairy science}, 5(3):249--258, 1922.

\bibitem{qian1999momentum}
Ning Qian.
\newblock On the momentum term in gradient descent learning algorithms.
\newblock {\em Neural networks}, 12(1):145--151, 1999.

\bibitem{qiao2019individual}
Yongliang Qiao, Daobilige Su, He Kong, Salah Sukkarieh, Sabrina Lomax, and
  Cameron Clark.
\newblock Individual cattle identification using a deep learning based
  framework.
\newblock {\em IFAC-PapersOnLine}, 52(30):318--323, 2019.

\bibitem{reynolds2009gaussian}
Douglas~A Reynolds.
\newblock Gaussian mixture models.
\newblock {\em Encyclopedia of biometrics}, 741:659--663, 2009.

\bibitem{robbins1951stochastic}
Herbert Robbins and Sutton Monro.
\newblock A stochastic approximation method.
\newblock {\em The annals of mathematical statistics}, pages 400--407, 1951.

\bibitem{shanahan2009framework}
C Shanahan, B Kernan, G Ayalew, K McDonnell, F Butler, and S Ward.
\newblock A framework for beef traceability from farm to slaughter using global
  standards: an irish perspective.
\newblock {\em Computers and electronics in agriculture}, 66(1):62--69, 2009.

\bibitem{stephen1990perceptron}
I Stephen.
\newblock Perceptron-based learning algorithms.
\newblock {\em IEEE Transactions on neural networks}, 50(2):179, 1990.

\bibitem{tadesse2003milk}
Million Tadesse and Tadelle Dessie.
\newblock Milk production performance of zebu, holstein friesian and their
  crosses in ethiopia.
\newblock {\em Livestock Research for Rural Development}, 15(3):1--9, 2003.

\bibitem{tharwat2014cattle}
Alaa Tharwat, Tarek Gaber, Aboul~Ella Hassanien, Hasssan~A Hassanien, and
  Mohamed~F Tolba.
\newblock Cattle identification using muzzle print images based on texture
  features approach.
\newblock In {\em Proceedings of the Fifth International Conference on
  Innovations in Bio-Inspired Computing and Applications IBICA 2014}, pages
  217--227. Springer, 2014.

\bibitem{van2008visualizing}
Laurens~JP van~der Maaten and Geoffrey~E Hinton.
\newblock Visualizing high-dimensional data using t-sne.
\newblock {\em Journal of machine learning research}, 9(nov):2579--2605, 2008.

\bibitem{vidal2021perspectives}
Maxime Vidal, Nathan Wolf, Beth Rosenberg, Bradley~P Harris, and Alexander
  Mathis.
\newblock Perspectives on individual animal identification from biology and
  computer vision.
\newblock {\em arXiv preprint arXiv:2103.00560}, 2021.

\end{thebibliography}
}
\end{document}